\documentclass[runningheads]{llncs}
\usepackage{graphicx}
\usepackage{amsmath,amssymb} % define this before the line numbering.
\usepackage{color}

\usepackage{subfigure}
\usepackage{tabularx}
\usepackage{rotating}
\usepackage{diagbox}
\makeatletter
\renewcommand{\@thesubfigure}{\hskip\subfiglabelskip}
\makeatother
\usepackage{multirow}

\begin{document}
\pagestyle{headings}
\mainmatter

\def\ACCV20SubNumber{***}  % Insert your submission number here

%===========================================================
\title{Adversarial Image Composition with Auxiliary Illumination} % Replace with your title
\titlerunning{Adversarial Image Composition}
\author{Fangneng Zhan\inst{1,2}\orcidID{0000-0003-1502-6847} \and
Shijian Lu\inst{1}\orcidID{0000-0002-6766-2506} \and
Changgong Zhang\inst{2} \and Feiying Ma\inst{2} \and Xuansong Xie\inst{2}
}

\authorrunning{Fangneng Zhan et al.}

\institute{Nanyang Technological University \and Alibaba DAMO Academy \\
\email{\{fnzhan,shijian.lu\}@ntu.edu.sg,\{changgong.zcg,feiying.mfy\}@alibaba-inc.com, xingtong.xxs@taobao.com}}

\maketitle

%===========================================================
\begin{abstract}
Dealing with the inconsistency between a foreground object and a background image is a challenging task in high-fidelity image composition. State-of-the-art methods strive to harmonize the composed image by adapting the style of foreground objects to be compatible with the background image, whereas the potential shadow of foreground objects within the composed image which is critical to the composition realism is largely neglected. In this paper, we propose an Adversarial Image Composition Net (AIC-Net) that achieves realistic image composition by considering potential shadows that the foreground object projects in the composed image. A novel branched generation mechanism is proposed, which disentangles the generation of shadows and the transfer of foreground styles for optimal accomplishment of the two tasks simultaneously. A differentiable spatial transformation module is designed which bridges the local harmonization and the global harmonization to achieve their joint optimization effectively. Extensive experiments on pedestrian and car composition tasks show that the proposed AIC-Net achieves superior composition performance qualitatively and quantitatively.
\end{abstract}

%===========================================================
\section{Introduction}
With the advances of deep neural networks (DNNs), image composition has been attracting increasing attention as a typical approach for image synthesis, augmented reality, etc. With a foreground object and a background image, a direct combination tends to introduce unrealistic artifacts in the composed image with various inconsistency in colors, illuminations, texture, etc. Several methods have been proposed for harmonizing the foreground object and the background image, including DNN-based image translation \cite{tsai2017dih}, local statistic transfer \cite{luan2018dph}, etc. On the other hand, existing methods typically simplify the harmonization by solving the image consistency within the foreground object region only but neglecting that realistic image composition goes far beyond that.

Specifically, a typical visual effect that is often associated with objects in scenes is their projected shadows due to the 3D nature of the world. While embedding a foreground object into a background image for image composition, the shadow effects of the foreground object are critical which can significantly improve the realism of the composed image if handled properly. Due to various constraints in 2D images, such as complicated illuminations, losing of 3D information, etc., traditional computer vision methods are highly susceptible to failure for the task of high-fidelity shadow generation.

\begin{figure}[t]
\centering
\includegraphics[width=1.0\linewidth]{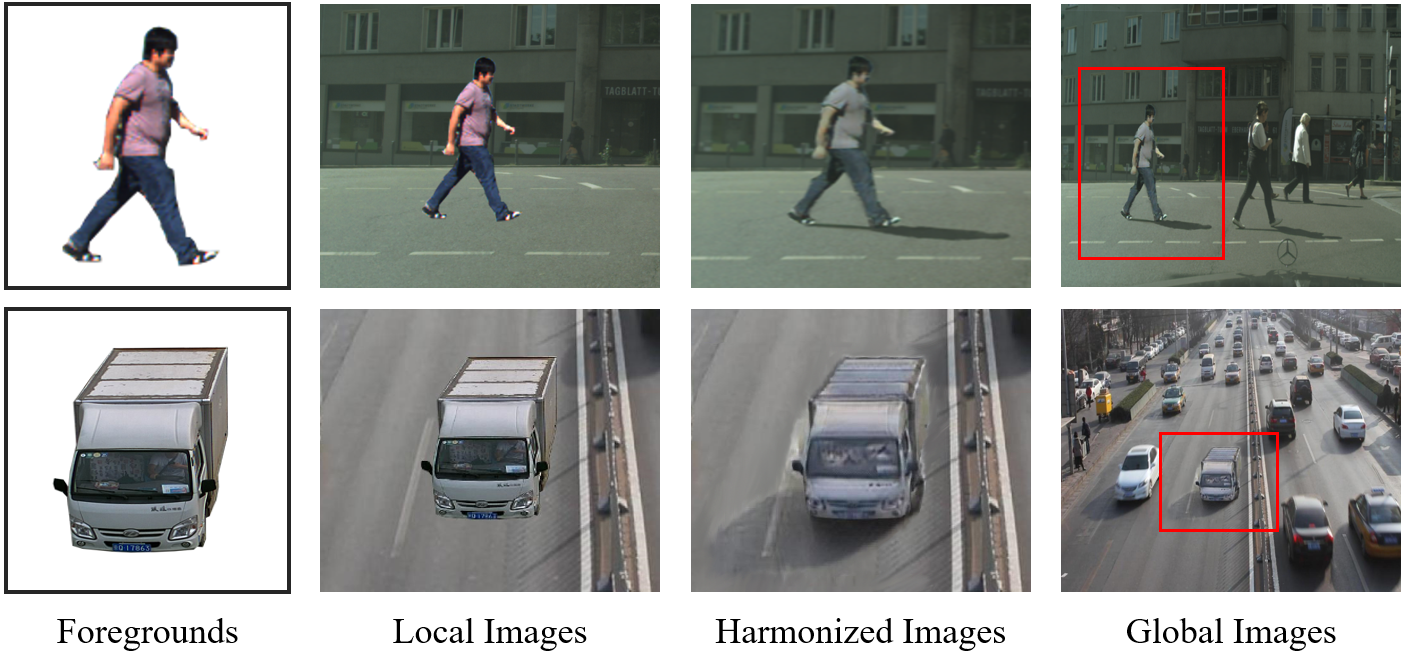}
\caption{
Given foreground objects and directly composed local images as shows in columns 1 and 2, the proposed AIC-Net disentangles the shadow generation and style transfer to produce realistic local harmonization as shown in column 3. With our spatial transformation module, the locally harmonized images in column 3 can be adapted and composed into the globally harmonized images as shown in column 4.
} 
\label{im_intro}
\end{figure}

In this work, we design a novel Adversarial Image Composition Net (AIC-Net) that aims to achieve realistic image composition with adaptive shadow effects as illustrated in Fig. \ref{im_intro}. AIC-Net adopts generative adversarial ideas to harmonize the foreground object within the background image and generates realistic shadow effects by learning from real shadows in natural images. It consists of local harmonization and global harmonization which are inter-connected and can be optimized jointly. We design a novel branched generation mechanism for high-fidelity local harmonization which disentangles the shadow generation and the style transfer of the foreground object for optimal accomplishment of two tasks simultaneously. An auxiliary illumination model is introduced to infer the illumination conditions of the background image such as lighting directions which help generate realistic shadows of the foreground object greatly. In addition, a differentiable spatial transformer is introduced to bridge the local region and the global background for joint optimization of the local and global harmonization. Two discriminators are employed where a local discriminator guides to learn style transfer and shadow generation of the foreground object and a global discriminator guide to learn global image harmonization.

The contributions of this work are three folds. First, we design an innovative AIC-Net that is capable of generating high-fidelity shadow of the foreground object adaptively. Second, we design a novel branched generation mechanism that disentangles the style transfer and shadow generation of the foreground object and accomplishes the two tasks optimally and simultaneously. Third, we design spatial transformer module (STM) that achieves joint optimization of local harmonization and global harmonization effectively.

\section{Related Work}

\subsection{Image Composition \& Harmonization}

Image composition aims to generate new images by embedding foreground objects into background images \cite{wu2017gpgan,lin2018stgan,zhan2018ver,zhan2019acgan,zhan2019gadan,zhan2019scene,zhan2019spatial,zhan2019esir,zhan2020towards,zhan2019spatial,liu2020arshadowgan}.
GP-GAN \cite{wu2017gpgan} composes high-resolution images by leveraging Poisson blending \cite{perez2003}.
% GP-GAN \cite{wu2017gpgan} synthesizes high resolution images by  Poisson blending \cite{perez2003}. 
\cite{lin2018stgan} presents a spatial transformer GAN by inserting STNs into the generator.
\cite{azadi2018comgan} describes a Compositional GAN that introduces a self-consistent composition-decomposition network. 
\cite{zhu2015} proposes a model to distinguish natural photographs from automatically generated composite images.
\cite{chen2019} proposes a generative adversarial network (GAN) architecture for automatic image compositing with consideration of geometric, color and boundary consistency.

Image harmonization deals with the inconsistency between a foreground region and a background image for composition. Several methods have been proposed to adapt the appearance of the foreground object for image harmonization. For example, \cite{sunkavalli2010multiscale} matches contrast, texture, and blur by manipulating the scales of a pyramid decomposition of images. \cite{xue2012} studies image statistics that determine the realism of the composite images. \cite{tao2013} proposes a gradient domain to achieve composition which can prevent color bleeding without changing the boundary location.
A number of deep-learning based image harmonization methods have also proposed in recent years. For example, \cite{tsai2016} utilizes visual semantics to guide the process of replacement to generate realistic, artifact-free images with diverse styles. \cite{luan2018dph} achieves both spatial and inter-scale statistical consistency by carefully determining the local statistics to be transferred. \cite{tsai2017dih} proposes an end-to-end deep convolutional neural network for image harmonization, which can capture both the context and semantic information of the composite images during harmonization. \cite{chen2019} combines a transformation network, a refinement network, a segmentation network and a pair of discriminator networks to achieve image composition via adversarial learning. Besides image harmonization, some work \cite{lalonde1,zhu2015} focuses on the realism assessment of the composed images.
\cite{cong2020dovenet} handles image harmonization through domain verification discriminator with the insight that the foreground needs to be translated to the same domain as background.

Illumination estimation is a classic computer vision problem and it is critical for achieving realistic lighting and shadows in image harmonization. Several deep learning based methods have been proposed to recover illumination from 2-dimensional RGB images \cite{Lalonde2012,gardner2017,geoffroy2017,gardner2019,garon2019,zhan2020emlight}.

\subsection{Generative Adversarial Networks}

GANs \cite{goodfellow2014gan} have achieved great success in image generation from either existing images or random noises. Instead of manually selecting features and parameters, GAN generators learn an optimal mapping from random noise or existing images to synthesized images, while GAN discriminators differentiate the synthesized images from real ones via adversarial learning. Several GAN-based image synthesis methods have been reported in recent years. \cite{denton2015lapgan} introduces Laplacian pyramids that improve the quality of GAN-synthesized images greatly. \cite{lee2018context} proposes an end-to-end trainable network for inserting an object instance mask into the semantic label map of an image. 
Other systems attempt to 
% synthesize realistic images by stacking a pair of generators \cite{zhang2017stackgan,zhang2018stackgan++}, 
learning more reasonable potential features \cite{chen2016infogan}, 
exploring new training approach \cite{arjovsky2017wgan}, 
visualizing GANs at the unit, object and scene level \cite{bau2019},
etc. 
To expand the applicability, 
\cite{liu2020arshadowgan} generate shadow for augmented reality by modeling the mapping relation between the virtual object shadow and the real-world.
% LR-GAN \cite{jwyang2017lrgan} synthesizes images by introducing spatial transformer networks (STNs). 
% \cite{yao2019,zhu2018von} study GAN-based 3D manipulation, etc.

GAN-based image-to-image translation has been widely explored due to its wide applicability. CycleGAN \cite{zhu2017cyclegan} proposes a cycle-consistent adversarial network for realistic image-to-image translation. \cite{isola2017pixel2pixel} investigates conditional adversarial networks for image-to-image translation. 
\cite{shrivastava2017simgan} adopts unsupervised learning to improve the realism of synthetic images using unlabelled real data. 
% \cite{huang2018munit} decomposes the image representation into a content code and a style code and recombine them to achieve multi-modal image-to-image translation. 
Other GANs \cite{zhu2017toward,azadi2018mcgan,park2019spade,liu2019funit} also achieve great performance in image-to-image translation.

\section{Proposed Method}
The proposed AIC-Net consists of local harmonization and global harmonization which are inter-connected and can be optimized jointly as illustrated in Fig. \ref{im_stru}. Detailed network structures and training strategies will be presented in the following subsections.

\subsection{Local Harmonization}

\begin{figure*}[t]
\centering
\includegraphics[width=1.0\linewidth]{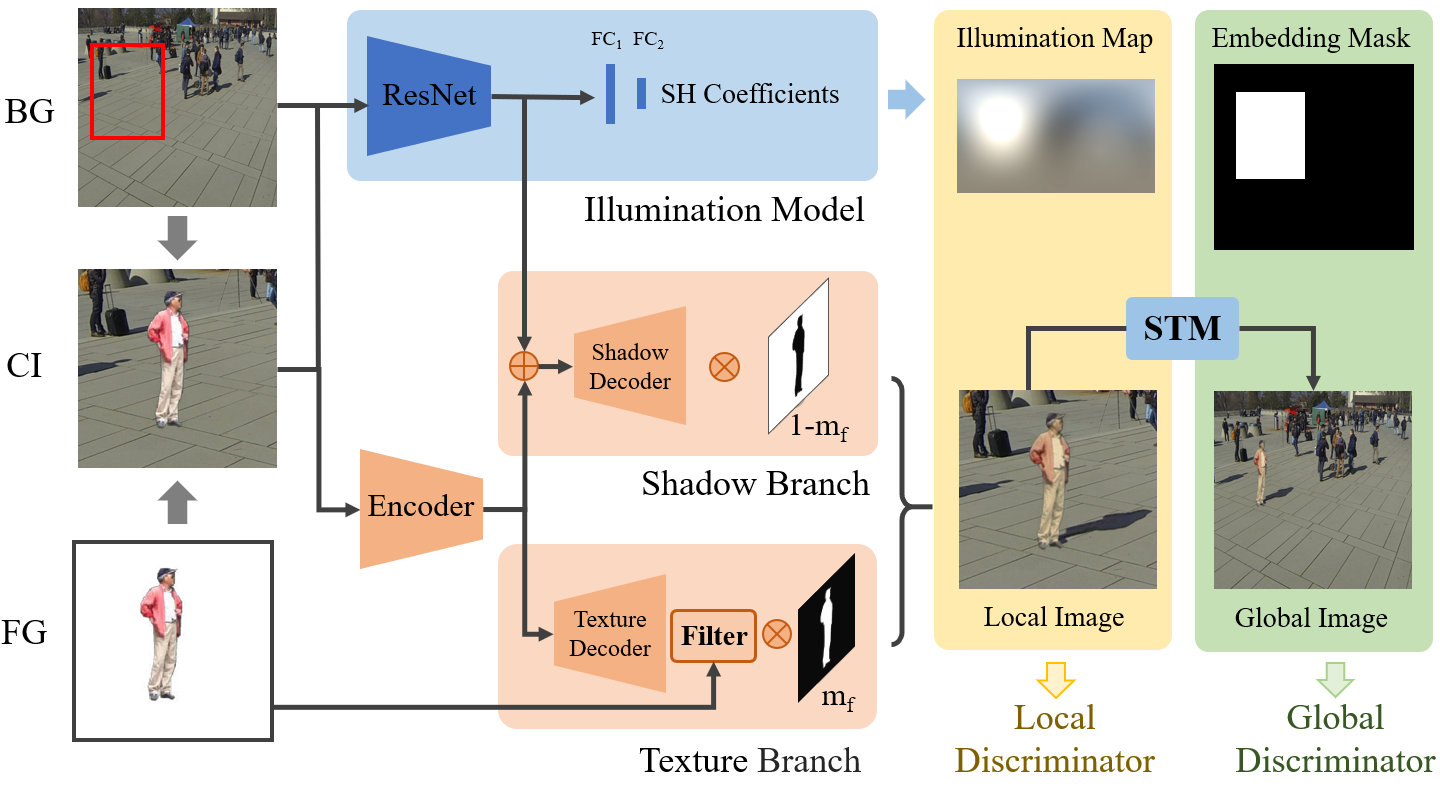}
\caption{The structure of the proposed AIC-Net: With a background image \textit{BG} (with specified composition regions) and foreground object \textit{FG} to be composed, a directly composed images \textit{CI} can be derived directly. A shadow branch and a texture branch will learn to generate shadow and transfer the style of \textit{CI} which produces a harmonized \textit{Local Image}. A spatial transformer module (STM) then learns to compose the \textit{Local Image} with the background image to produce a \textit{Global Image}. An \textit{Illumination Model} will predict SH coefficients which can be used to reconstruct an \textit{Illumination Map} (`FC' denote fully connected layer). The \textit{Local Image} will be concatenated with the \textit{Illumination Map} to be distinguished by the \textit{Local Discriminator}. The \textit{Global Image} will be concatenated with an \textit{Embedding Mask} to be distinguished by the \textit{Global Discriminator}. 
}
\label{im_stru}
\end{figure*}

The local harmonization is achieved by a local discriminator and a branched generation network which consists of a shadow branch and a texture branch as illustrated in Fig. \ref{im_stru}. Given a background image (BG), a meaningful region within it is first selected for composition with a foreground object (FG). The selection of the meaningful region will be described in more details in \textit{Experiment Setting}. We thus obtain a directly composed image (CI) $x$ by embedding the foreground object into the selected region in the background image, as well as a foreground mask $m_{f}$ by setting the foreground object pixels to 1 and the rest pixels in $x$ to 0 as illustrated in Fig. \ref{im_stru}

In local harmonization, the generator aims to transfer the style of the composed image and generate realistic object shadows concurrently. Although the two processes can be jointly handled through image-to-image translation, they have very different translation objectives and learn with distinct image-to-image mapping mechanisms. The concurrent learning of both degrades the translation severely as illustrated in `W/o Branched Generation' in Fig. \ref{im_method}. We design a novel branched generation strategy to disentangle style transfer and shadow generation for optimal image harmonization. As shown in Fig. \ref{im_stru}, the generator has two branches that share the same Encoder. With a background image and a composed image as the input, the decoder in shadow branch strives to generate shadows in the composed image while the decoder in the texture branch strives to transfer the style of the composed image to be compatible with the background image.

\subsubsection{Shadow Branch}

\begin{figure}[ht]
\centering
\includegraphics[width=1.0\linewidth]{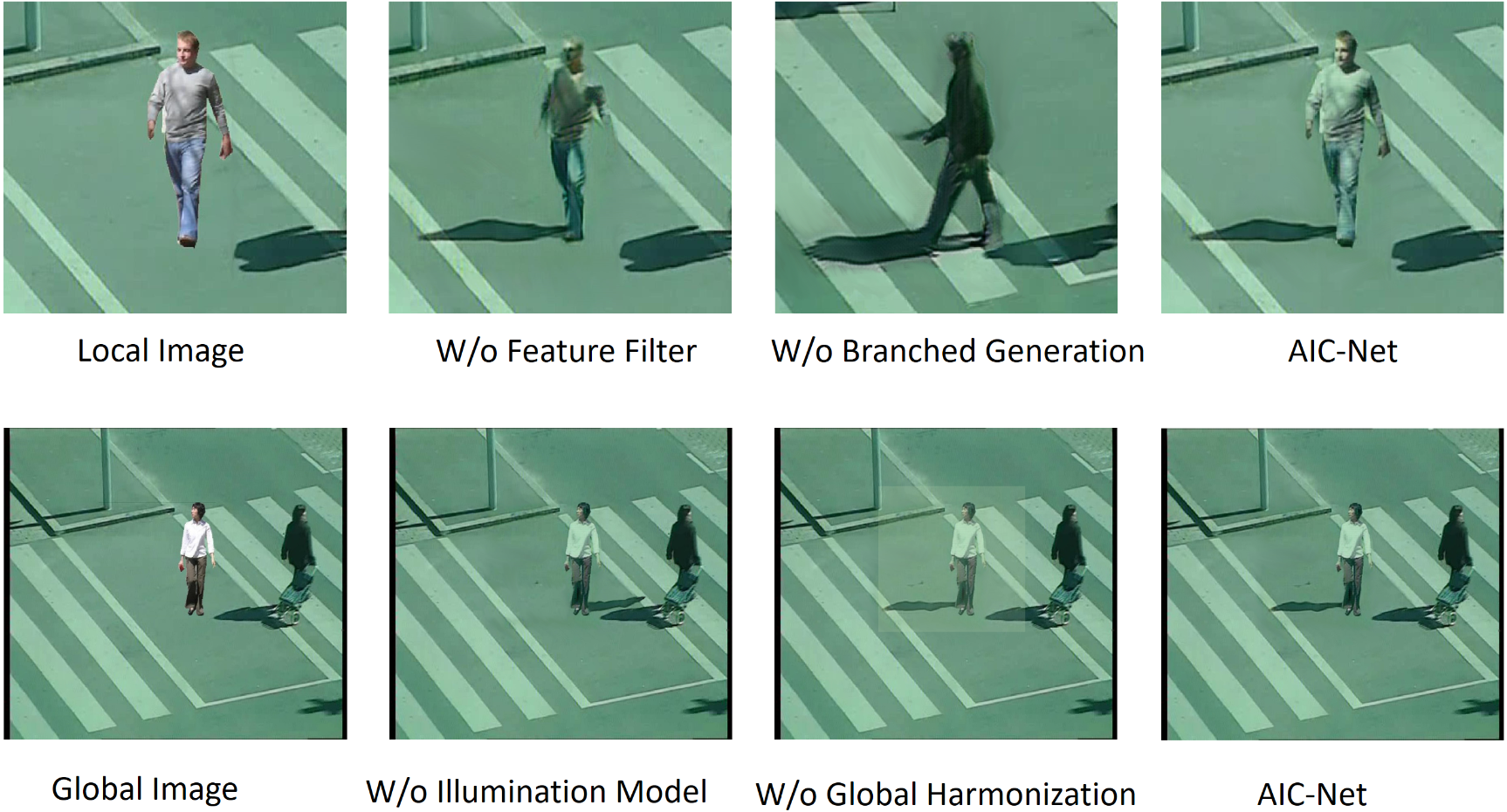}
\caption{Illustration of the proposed guided feature filter, branched generation, illumination model and global harmonization: The introduction of the branched generation and guided feature filter helps to suppress various artifacts and generate realistic shadows and foreground object in the local image. The illumination model helps to generate shadow with high-fidelity shadow styles and directions. The global harmonization helps to suppress artifacts and style discrepancy between the local and global images.
}
\label{im_method}
\end{figure}

To avoid synthesizing shadows with false directions as shown in `W/o Illumination Model' in Fig. \ref{im_method}, we need to infer the global illumination information from the background images. In this work, we adopt a deep learning based method for illumination estimation. Specially, the ground truth of illumination map which is a panorama is represented by spherical harmonics (SH) coefficients \cite{shlight}, then a model with ResNet \cite{resnet} as the backbone is adopt to regress the SH coefficients from the background images with limited field of view. With the predicted SH coefficients, an illumination map $M_{s}$ can reconstructed as follows:
\begin{equation}
M(s) =\sum_{i=1}^{M}SH_{i}*y_{i}(s)
\end{equation}
where $s$ is the spatial direction, $M$ is the number of SH coefficients, $y_{i}$ and $SH_{i}$ are the i-th SH basis function and SH coefficients, respectively.
We pre-train the illumination model on the Laval Indoor dataset \cite{gardner2017} and \cite{cheng2018shlight} and fix the model weights while training AIC-Net.

The illumination model is able to infer the illumination from cues such as shadow existed in the background. Thus, the feature extracted by the illumination model contains rich information of shadow direction, size and intensity.
In the shadow branch as shown in Fig. \ref{im_stru}, we concatenate the feature extracted by the illumination model and the feature extracted by Encoder to integrate the shadow and illumination information into the shadow generation. The shadow decoder then strives to generate shadows with realistic shadow directions and styles in the local images. A binary mask (1-$m_{f}$) is applied to the output for integration with the texture branch. As the shadow branch may translate other regions beyond the shadow areas undesirably, we feed real images into the shadow branch to keep the non-shadow areas be the same with the input. The shadow branch will thus concentrate on the translation of potential shadow areas for realistic shadow generation.
To drive the effective learning of shadow branch, the reconstructed illumination map will be combined with the locally harmonized image to be distinguished by the local discriminator.

\begin{figure*}[t]
\centering
\includegraphics[width=1.0\linewidth]{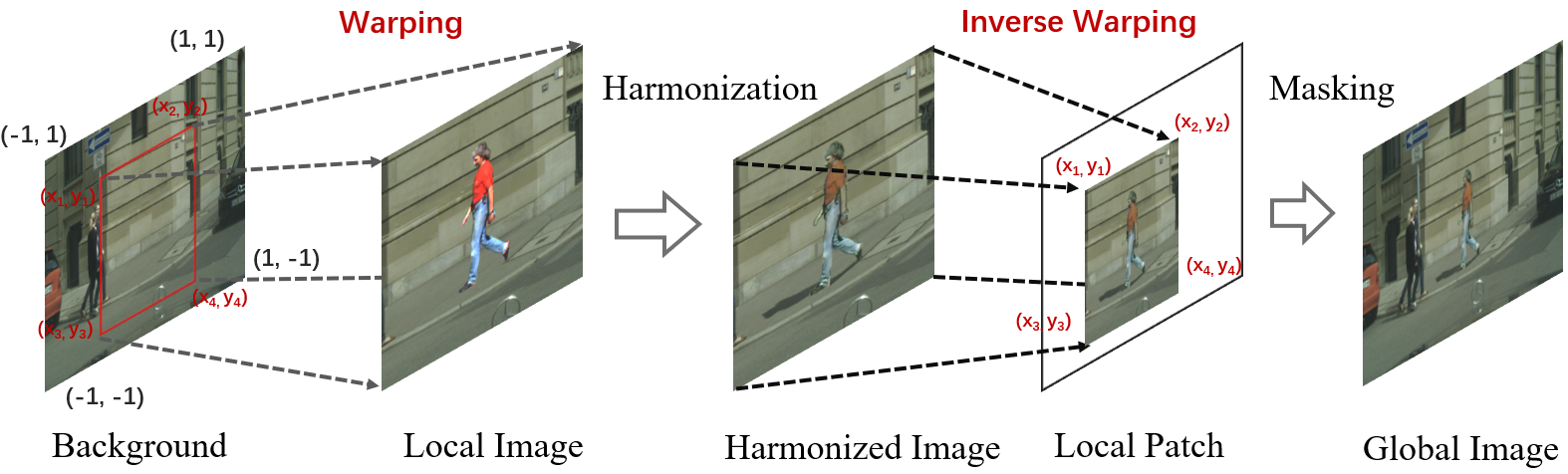}
\caption{
Global harmonization: With a selected region in the background image, a homography matrix $H$ can be estimated to transform the selected region to the local image. After the local harmonization, the harmonized local image can be transformed back to the background image by an inverse warping $H^{-1}$. A masking operation is followed to obtain the global image.
}
\label{im_spatial}
\end{figure*}

\subsubsection{Texture Branch}

The texture branch aims to translate the texture of the foreground region to be harmonious with the background image. As a normal GAN generator tends to over-translate the foreground region which often degrades the foreground features as illustrated in `W/o Feature Filter' in Fig. \ref{im_method}, we introduce a guided feature filter to preserve the `content features’ while performing style transfer. Different from the guided image filter \cite{he2013} which is conducted in image level, guided feature filter is conducted in the feature space to achieve flexible style transfer. With the foreground object as the content guidance (C) and the feature decoded by the Texture Decoder as the style guidance (S), an output feature (T) with transferred style but preserved contents can be obtained through a linear model as follows:
\begin{equation}
T_{i} = a_{k}C_{i}+b_{k}, \forall i \in \omega_{k}
\end{equation}
where $i$ is a pixel index and $\omega_{k}$ is a local square window centered at pixel $k$, $a_{k}$ and $b_{k}$ are linear coefficients to be optimized.
The minimization of the reconstruction error between $C$ and $T$ can be achieved by the following cost function:
\begin{equation}
E(a_{k},b_{k})=\sum_{i \in \omega_{k}}((a_{k}S_{i}+b_{k}-C_{i})^{2}+\epsilon a_{k}^2)
\end{equation}
where $\epsilon$ is a regularization parameter that prevents $a_{k}$ from being too large. It can be solved via linear regression:
\begin{equation}
a_{k}=\frac{
\frac{1}{\left| \omega \right|} 
\sum_{i \in \omega_{k}} S_{i}-\mu_{k}\overline{C}_{k}}
{\overline{\sigma}_{k}+\epsilon}
\end{equation}
\begin{equation}
b_{k}=\overline{R}_{k}-a_{k}\mu_{k}
\end{equation}
where $\mu_{k}$ and $\sigma_{k}^2$ are the mean and variance of $S$ in $\omega_{k}$, $\left| \omega \right|$ is the number of pixels in $\omega_{k}$, and $\overline{C}_{k}=\frac{1}{\left| \omega \right|} \sum_{i \in \omega_{k}}C_{i}$ is the mean of $C$ in $\omega_{k}$. Then we can obtain the transferred feature $T$.
A following convolution layer will finally generate images with transferred style. A binary mask $m_{f}$ is applied to the output for further integration with the shadow branch. By denoting the outputs of the shadow branch and the texture branch by $x_{s}$ and $x_{t}$, the locally harmonized image $x_{h}$ can be obtained as follows:
\begin{equation}
x_{h} = x_{t} * m_{f} + x_{s} *(1-m_{f})
\end{equation}

\subsection{Global Harmonization}
% The harmonized local region should be consistent with global image regarding the image style especially the boundary of the local region in the global image, otherwise the global image will tend to present clearly artifacts as shown in `W/o Global Harmonization' of Fig. \ref{im_method}.
% A global discriminator is thus adopted to ensure the global consistency.
% To indicate the embedding region in the global image, an embedding mask with the selected region as 1 and other region as 0 is concatenated with the global image as the input of global discriminator.
% With real background image as the training reference, the global discriminator learns to distinguish the pair of Global Image and Embedding mask, driving the generator to generate consistent local image with the global image.
% To bridge the local image and global image to achieve joint optimization, a spatial transformer module (STM) is designed in the network.

The harmonized local region should be consistent with global image regarding the image style especially the boundary of the local region in the global image, otherwise the global image will tend to present clearly artifacts as shown in `W/o Global Harmonization' of Fig. \ref{im_method} which should be addressed as well for high-fidelity global harmonization. We adopt a global discriminator to ensure the consistency in global harmonization. With real background image as the training reference, the global discriminator learns to distinguish the harmonized global image (Fake Global in Fig. \ref{im_stru}) and real global images (Real Global in Fig. \ref{im_stru}), driving the generator for generating harmonious shadows and an attention network for eliminating the style discrepancy and boundary artifacts. A spatial transformer module (STM) is designed to bridge the local image and global image to achieve joint optimization.

\textbf{Spatial Transformer Module:}
The spatial transformer is a differentiable module that applies a homography to the local image as illustrated in Fig. \ref{im_spatial}. We denote the coordinates of the four vertices of the background image by [(-1, 1), (1, 1), (1, -1), (-1, -1)], and the coordinate of the four vertices of the selected region by [(x1, y1), (x2, y2), (x3, y3), (x4, y4)] ($-1<x_{i}<1, -1<y_{i}<1, i=1,2,3,4$). With four paired vertices, the homograhy $H$ can be computed which applied STM to warp the selected region to the local image. Once the local image is harmonized, it can be warped back to the local region of the background image by using the inverse homography $H_{-1}$ as illustrated in Fig. \ref{im_spatial}. A warping mask with the selected region as 1 and other regions as 0 can be obtained simultaneously. A globally harmonized image can thus be obtained by masking the local patch into the background image. Since the STM and the masking operation are both differentiable, the local harmonization and the global harmonization can be bridged for end-to-end joint optimization.

\subsection{Adversarial Training}
With a GAN-based structure, the network is trained through adversarial learning between a generator and two discriminators. For clarity purpose, we denote the \textit{BG}, \textit{CI} in Fig. \ref{im_stru} as $X$ and $x$, the generator and the two branches \textit{Shadow Branch} and \textit{Texture Branch} as $G$, $G_{s}$ and $G_{t}$, \textbf{Global Discriminator} and \textbf{Local Discriminator} as $D_{G}$ and $D_{L}$.
The foreground mask is denoted as $m_{f}$, the real global and local images as $Y$ and $y$. The \textbf{Embedding Mask} in real global images and harmonized global images as $m_{Y}$ and $m_{X}$.

For the local harmonization, the shadow and texture branches strive to generate shadow and translate foreground that are denoted by $G_{s}(X, x)$ and $G_{t}(X, x)$, respectively. The local image $x_{h}$ can thus be denoted by:
\begin{equation}
\begin{split}
x_{h} = G_{s}(X, x) * (1-m_{f}) + G_{t}(X, x) * m_{f}
\end{split}
\end{equation}
The local discriminator tries to distinguish the local image while the generator tries to mislead the local discriminator.
As we adopt Wasserstein GAN \cite{arjovsky2017wgan} objective for training, the local adversarial loss of the generator and local discriminator can be defined by:
\begin{equation}
\begin{split}
L_{D_{L}} = E[D_{L}(x_{h})] - E[D_{L}(y)]
\end{split}
\end{equation}
\begin{equation}
\begin{split}
L_{G_{L}} = -E[D_{L}(x_{h})]
\end{split}
\end{equation}
To ensure the irrelevant region keep unchanged and preserve more details of the original image, an identity loss is defined as follows:
\begin{equation}
L_{S_{idt}} = E[G(Y, y) - y]
\end{equation}

For the global harmonization, the global discriminator will distinguish the pair of global image $X_{h}$ and embedding mask $m_{X}$ which drives the generator to generate consistent texture and suppress artifacts around the local image boundary. The global adversarial loss of the generator and the global discriminator can thus be defined by:
\begin{equation}
\begin{split}
L_{D_{G}} = E[D_{G}(X_{h}, m_{X})] - E[D_{G}(Y, m_{Y})]
\end{split}
\end{equation}
\begin{equation}
\begin{split}
L_{G_{G}} = -E[D_{G}(X_{h}, m_{X})]
\end{split}
\end{equation}

The overall objective function for the generator is:
\begin{equation}
L_{G} = L_{G_{L}} + \lambda_{G} L_{G_{G}} + \lambda_{G_{idt}} L_{S_{idt}}
\end{equation}
where $\lambda_{G}, \lambda_{G_{idt}}$ denote weights of global adversarial loss and generator identity loss respectively.
The overall objective function for the discriminators is:
\begin{equation}
L_{D} = L_{D_{L}} + \lambda_{D_{G}} L_{D_{G}}
\end{equation}
where $\lambda_{D_{G}}$ denotes the weight of global adversarial loss for the discriminator.
$L_{D}$ and $L_{G}$ drive the learning of the whole model alternately.

\section{Experiment}

\subsection{Dataset}

\textbf{PRID 2011} \cite{prid} was created to evaluate person re-identification methods. The dataset consists of images extracted from multiple person trajectories recorded from two surveillance cameras. 

\textbf{WILDTRACK} \cite{wildtrack} is multi-camera detection and tracking dataset of pedestrians in the wild, where the cameras’ fields of view has large part overlap. 

\textbf{Penn-Fudan} \cite{penn-fudan} is an image database that was used for pedestrian detection evaluation as reported in [1]. The images are taken from scenes around campus and urban streets. The objects we are interested in these images are pedestrians. Each image will have at least one pedestrian.

\textbf{UA-DETRAC} \cite{detrac} dataset consists of 100 challenging video sequences captured from over 140 thousand frames of real-world traffic scenes. More than 1.2 million labeled bounding boxes of vehicles with information of occlusion, weather, vehicle category and truncation are annotated.

\textbf{Cityscapes} \cite{cityscapes} dataset focuses on semantic understanding of urban street scenes. It has street scenes from 50 cities and provides semantic segmentation of 30 classes. 3475 images are annotated with bounding boxes of persons. 

\textbf{ShapeNet} \cite{shapenet} is a large repository of 3D object models which index more than 3 million models of which 220 thousand are classified into 3,135 categories.

\subsection{Experiment Setting}
In the pedestrian harmonization experiment, 2000 background images are collected from \cite{prid} and \cite{wildtrack} and 500 pedestrians are cropped from \cite{penn-fudan} as the foreground objects. In the car harmonization experiment, we collect the background images from \cite{detrac} and \cite{cityscapes} which contain images captured under different illumination conditions and viewpoints. Random horizontal flipping is applied to the background images to augment the training set. For the foreground images, 1000 3D car models from \cite{shapenet} are utilized to render 3000 car images in different views.

The local region to compose with the foreground object is derived by a simple mechanism. As the bounding boxes of the foreground objects are know, we first select a box region around an existed bounding box and the one with no overlap with other bounding boxes is selected as the local region.

The proposed method is implemented using the Tensorflow framework.  The optimizer is Adadelta which employs adaptive learning rate. The model is trained in 100 thousand iterations with a batch size of 4. In addition, the network is trained on a workstation with an Intel Core i7-7700K CPU and a NVIDIA GeForce GTX 1080 Ti graphics card with 12GB memory. All input images are resized to $256 \times 256$ in training.

\renewcommand\arraystretch{1.25}
\begin{table*}[ht]
\small
\caption{Comparison of harmonization methods with different metrics: `Car' and `Ped' denote the results of car and pedestrian synthesis respectively. `Local' and `Global' denote harmonized local and global images. `MS' metric denotes the manipulation scores. For AMT user study, the number in each cell is the percentage of the harmonized images that are deemed as the most realistic by Turkers.}
\renewcommand\tabcolsep{3.8pt}
\centering 
\begin{tabular}{l|p{0.75cm}<{\centering}p{0.75cm}<{\centering}|cc|p{0.6cm}p{0.6cm}|cc|cc} \hline
\multirow{1}{*} & 
\multicolumn{2}{c|}{\textbf{CycleGAN}} & 
\multicolumn{2}{c|}{\textbf{DeepIH}} & 
\multicolumn{2}{c|}{\textbf{DeepPH}} &
\multicolumn{2}{c|}{\textbf{AdaIN}} &
\multicolumn{2}{c}{\textbf{AIC-Net}} \\
\cline{2-11}
\textbf{Metrics} & Car & Ped & Car & Ped & Car & Ped & Car & Ped & Car & Ped \\\hline
\textbf{AMT(Local)}    & 8\% & 4\%   & 12\% & 11\%   & 20\% & 23\%   & 18\% & 17\%    & \textbf{42\%} & \textbf{45\%}   \\
\textbf{AMT(Global)}    & 3\% & 1\%   & 16\% & 15\%   & 25\% & 24\%   & 21\% & 22\%    & \textbf{41\%} & \textbf{42\%}   \\\hline
\textbf{FID(Local)} & 115.1 & 131.6   & 136.2 & 151.4   & 99.1 & 103.3   & 88.7 & 97.2    & \textbf{84.4} & \textbf{97.1}   \\
\textbf{FID(Global)} & 102.4 & 126.1   & 121.8 & 139.3   & 92.4 & 101.9   & 84.6 & 92.8    & \textbf{72.2} & \textbf{89.6} \\\hline
\textbf{MS(Local)}    & 0.69 & 0.70   & 0.75 & 0.74   & 0.72 & 0.73   & 0.69 & 0.71    & \textbf{0.63} & \textbf{0.64}   \\
\textbf{MS(Global)} & 0.71 & 0.70   & 0.73 & 0.76   & 0.73 & 0.74   & 0.70 & 0.74   & \textbf{0.62} & \textbf{0.62}   \\\hline
\end{tabular}
\label{tab_compare}
\end{table*}

\subsection{Experiment Analysis}
\subsubsection{Quantitative Analysis:}

The evaluation is performed through Frechet Inception Distance (FID) \cite{fid}, Manipulation Score (MS) \cite{chen2019} and Amazon Mechanical Turk (AMT) study.
For the evaluation with FID, the harmonized images are fed to the Inception network \cite{inception} trained on ImageNet \cite{imagenet}, and the features from the layer before the last fully-connected layer are used to calculate the Fréchet Inception Distance. 
The Manipulation Score \cite{chen2019} is generated by a state-of-the-art manipulation detection algorithm \cite{ms} where a higher score indicates a higher possibility that the image is manipulated.
For the AMT study which involves 10 users, we present a set of images (5 images by 5 methods) each time to the user to select the most realistic one. With 20 sets of presented images, an AMT score is derived by averaging the percentage of most realistic images as judged by the 10 users.
`Local' and `Global' denote the results of locally and globally harmonized images respectively.

Table \ref{tab_compare} compares the quality and realism of images processed by different harmonization methods. 
% As shown in Table \ref{tab_compare}, 
For the local harmonization, AIC-Net outperforms all compared methods with all metrics clearly as it harmonizes the foreground region and generates relevant shadow which improves the image realism significantly.
The using of guided filter helps to preserve texture of high quality, thus obtaining the best FID scores.
The spatial transformer module achieves the joint optimization of local and global harmonization, leading to best Manipulation score.
CycleGAN can also synthesize shadows but the foreground region tends to be translated undesirably as it does not disentangle the shadow generation and style transfer processed.
As DeepIH requires paired training images with semantic annotations, it performs terribly bad when paired training images are not available. As style transfer methods, DeepPH and AdaIN harmonize the foreground region successfully but all of them ignore the shadow generation completely. 
% The last two rows compare the realism of global images processed by different harmonization methods. 
% As Table \ref{tab_compare} shows, 
For the global harmonization, AIC-Net outperforms other methods by large margins as its joint optimization of local and global harmonization ensures consistent shadow and smooth local boundary. The lack of global harmonization leads to clear artifacts in the global image harmonized by CycleGAN. For DeepIH, DeepPH and AdaIN, there are not many artifacts but the lack of shadow and guided filter degrades the image realism and quality greatly.
Normally, the global harmonized images present better score than local harmonized images as only small part of global image needs to be harmonized.

\begin{figure*}[t]
\centering
\includegraphics[width=1.0\linewidth]{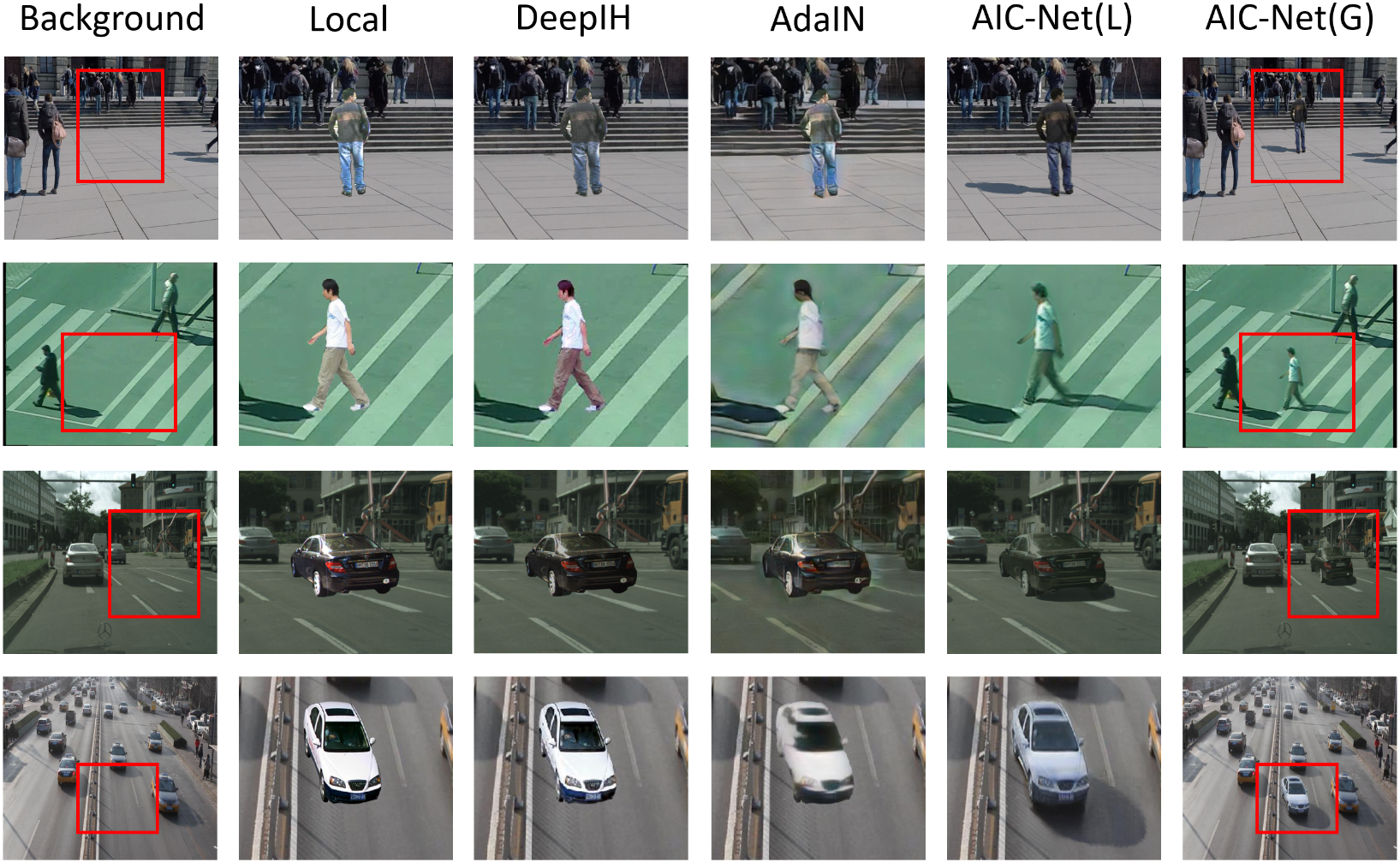}
\caption{
Comparing AIC-Net with other harmonization methods: Images in col. 1 are background images with selected regions. Col. 2 shows directly composed local images. AIC-Net (L) and AIC-Net (G) show images after local and global harmonization.
}
\label{im_comp}
\end{figure*}

\begin{figure*}[t]
\centering
\includegraphics[width=1.0\linewidth]{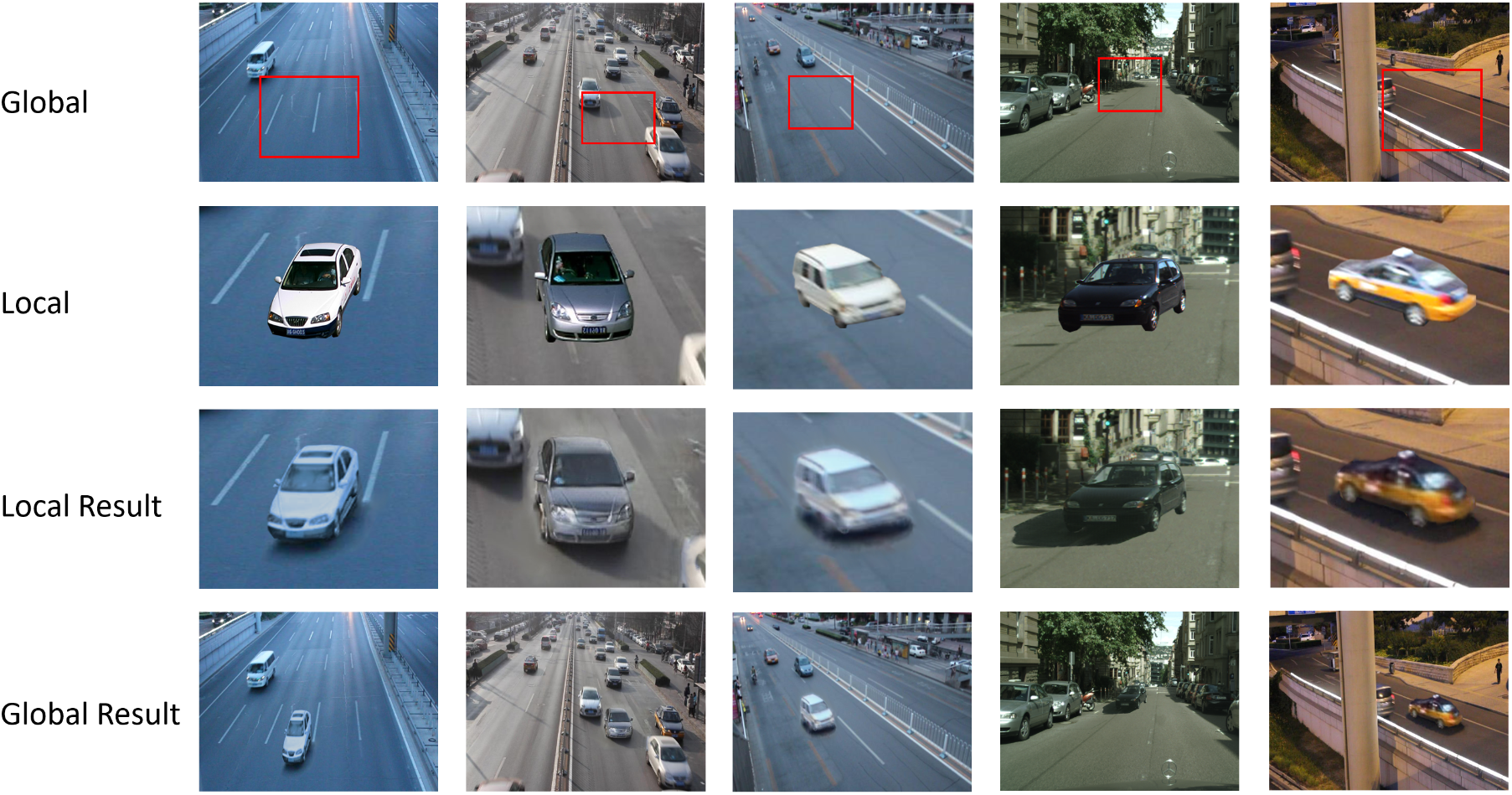}
\caption{Shadow generation under different conditions by AIC-Net: The first row shows the background image with the local region for foreground embedding. The second row show the direct composition. Rows 3-4 show the local harmonization and global harmonization by AIC-Net, respectively. AIC-Net can generate various shadows adaptively conditioned on illumination, viewpoint and weather of the background image, etc.
}
\label{im_cam}
\end{figure*}

\subsubsection{Qualitative Analysis:}

We compare the images harmonized by different methods in Fig. \ref{im_comp}, where AIC-Net (L) and AIC-Net (G) denote harmonization of local images and global images by our proposed AIC-Net. 
As Fig. \ref{im_comp} shows, 
% CycleGAN can generate shadows but the appearance of the foreground objects is degraded due to the concurrent style transfer and shadow generation. 
DeepIH \cite{tsai2017dih} can only achieve a slight translation and cannot generate any potential shadow, thus resulting in unrealistic harmonized images.
The branched generation in AIC-Net disentangles the style transfer and shadow generation, leading to much more realistic image harmonization. 
The global discriminator in AIC-Net works better by driving the generator to remove the boundary artifacts. AdaIN \cite{huang2017adain} transfer the style of the foreground region to be harmonious with the background, but they do not handle shadow generation which leads to unrealistic composition. Besides, AdaIN is unable to preserve the texture quality of harmonized image due to the missing of guided filter.

To validate that the proposed model is able to generate various shadow adaptively, we tested the trained model in different environments as shown in Fig. \ref{im_cam}.
The first two rows show the backgrounds and local images respectively, the last row shows locally harmonized images and globally harmonized images produced by our model.
As shown in the first three columns of Fig. \ref{im_cam}, the proposed model generates shadows conditioned on the existed shadow cases in the backgrounds.
% The size, strength and direction of the shadow are determined adaptively according
For background images without clearly shadow case as shown in the last two columns of Fig. \ref{im_cam}, the proposed model is still able to synthesize realistic shadows conditioned on the feature extracted by the illumination model.

\renewcommand\arraystretch{1.2}
\begin{table*}[ht]
\small
\caption{We conduct ablation study with three metrics (FID, AMT, MS) to evaluate the quality of harmonized local and global images. 
AIC-Net (WF), AIC-Net (WB), AIC-Net (WS) and AIC-Net(WI) denote AIC-Net without the guided feature filter, without branched generation mechanism, without spatial transformer module and without illumination model respectively. For AMT user study, the percentage numbers represent how often the harmonized images in each category are classified as \textit{most realistic} by Turkers. We average the FID, AMT and MS scores of car and pedestrian images as the final scores.}
\renewcommand\tabcolsep{3.8pt}
\centering 
\begin{tabular}{l|p{1.2cm}<{\centering}p{1.2cm}<{\centering}|p{1.2cm}<{\centering}p{1.2cm}<{\centering}|p{1.2cm}<{\centering}p{1.2cm}<{\centering}} \hline
\multirow{1}{*} & 
\multicolumn{2}{c|}{\textbf{FID}} & 
\multicolumn{2}{c|}{\textbf{AMT}} & 
\multicolumn{2}{c}{\textbf{MS}} 
% \multicolumn{2}{c}{\textbf{AIC-Net}} 
\\
\cline{2-7}
\textbf{Methods} & Local & Global & Local & Global & Local & Global \\\hline
\textbf{AIC-Net(WF)} & 115.5 & 106.1   & 14\% & 19\%   & 0.72 & 0.70   \\
\textbf{AIC-Net(WB)}    & 109.2 & 93.7  & 19\% & 20\%    & 0.70 & 0.69   \\
\textbf{AIC-Net(WS)}    & 98.5 & 92.1    & 21\% & 16\%    & 0.73 & 0.71   \\
\textbf{AIC-Net(WI)}    & 94.2 & 91.8    & 17\% & 18\%    & 0.67 & 0.65   \\
\textbf{AIC-Net}    & 90.8 & 81.8    & 29\% & 27\%    & 0.63 & \textbf{0.62}   \\
\hline
\end{tabular}
\label{tab_ablation}
\end{table*}

\subsubsection{Ablation Study:}

We conducted an ablation study to evaluate the effectiveness of different technical designs in AIC-Net. As shown in Table \ref{tab_ablation}, four variants of AIC-Net are designed including: 1) AIC-Net (WF) denotes AIC-Net without the guided feature filter; 2) AIC-Net (WB) denotes AIC-Net without using the branched generation mechanism; 3) AIC-Net (WS) denotes AIC-Net without incorporating the spatial transformer module; 4) AIC-Net(WI) denotes AIC-Net without illumination model.
% i.e. the local and global harmonization are conducted separately. 
The evaluations are performed with three metrics including FID, MS and AMT study that involves 10 users. For AMT user study, a set of images (5 generated by AIC-Net and its four variants) are presented to each of 10 users who will choose the most realistic image. We presented 25 sets of harmonized images and the AMT score is derived by averaging the percentages of the most realistic images as judged by the 10 users.
The scores of car and pedestrian experiments are averaged for each metric to obtain the final scores.

% As Table. \ref{tab_ablation} shows, AIC-Net achieves the lowest FID and it also outperforms AIC-Net(WF) and AIC-Net(WB) clearly, demonstrating the effectiveness of the proposed and guided filter.
 
As Table \ref{tab_ablation} shows, AIC-Net (WF) obtains clearly lower scores with all metrics as compared with the standard AIC-Net which shows the importance of guided feature filter in generating realistic image harmonization. AIC-Net (WB) also obtains lower scores than AIC-Net as the branched generation in AIC-Net disentangles the shadow generation and style transfer to produce more realistic local harmonization as shown in Fig. \ref{im_method}. For AIC-Net (WS), the scores for the global images are much lower than the other models, demonstrating the importance of the spatial transformer module which bridges the local and global harmonization to produce globally realistic harmonization. On the other hand, AIC-Net (WS) achieves similar scores which can be expected as both models are almost the same in local harmonization.
AIC-Net (WI) present clearly drop of AMT score compared with standard AIC-Net, indicating the significance of illumination model for the generation of realistic shadow.
% the AMT scores for the global images are much lower than the other three models, demonstrating the importance of the spatial transformer module which bridges the local and global harmonization to produce globally realistic harmonization.

\section{Conclusion}

This paper presents AIC-Net, an end-to-end trainable network that harmonizes foreground objects and background images with adaptive and realistic shadow effects.
% The AIC-Net is capable of harmonize the foreground region and generate adaptive shadows according to the background environments.
A branched generation mechanism is designed to disentangle style transfer and shadow generation to produce high-fidelity local harmonization. An auxiliary illumination model is collaborated in shadow branch to achieve adaptive shadow generation. A spatial transformer module is introduced to bridge the local and global harmonization for joint optimization.
The quantitative and qualitative experiments demonstrate the superior harmonization performance and wide adaptability of the proposed method.
We will continue to study AIC-Net for image harmonization with shadow generation within more complicated illumination environments.

%===========================================================
\bibliographystyle{splncs}
\bibliography{egbib}

%this would normally be the end of your paper, but you may also have an appendix
%within the given limit of number of pages
\end{document}